\documentclass{llncs}
\usepackage{times}
\usepackage{url}
\usepackage{latexsym}
\usepackage[small,bf]{caption} 
\usepackage{subcaption}

\usepackage{amsmath}
\usepackage{booktabs}
\usepackage{multirow}
\usepackage{graphicx}

\usepackage{hyperref}
\hypersetup{colorlinks,citecolor=black, linkcolor=black, urlcolor=black}
\usepackage{natbib}
	
\usepackage[nameinlink,capitalize]{cleveref}

\newcommand{\x}{\mathbf{x}}
\newcommand{\y}{\mathbf{y}}

\newcommand{\VTheta}{\mathbf{\Theta}}

\DeclareMathOperator*{\argmax}{arg\,max~}

\newcommand{\p}{\mathrm{p}}

\newcommand*{\regularbox}[1]{
	\setlength{\fboxsep}{-1\fboxrule}
	\setlength{\fboxrule}{0.3pt}
	\fbox{\hspace{1pt}\strut#1\hspace{1pt}}~
	\color{black}  }

\title{An Interactive Machine Translation Framework for Modernizing Historical Documents}

\author{Miguel Domingo \and Francisco Casacuberta}
\institute{PRHLT Research Center \\ Universitat Polit{\`e}cnica de Val{\`e}ncia \\ \email{midobal@prhlt.upv.es, fcn@prhlt.upv.es}}

\date{}

\begin{document}
\maketitle
\begin{abstract}
Due to the nature of human language, historical documents are hard to comprehend by contemporary people. This limits their accessibility to scholars specialized in the time period in which the documents were written. Modernization aims at breaking this language barrier by generating a new version of a historical document, written in the modern version of the document's original language. However, while it is able to increase the document's comprehension, modernization is still far from producing an error-free version. In this work, we propose a collaborative framework in which a scholar can work together with the machine to generate the new version. We tested our approach on a simulated environment, achieving significant reductions of the human effort needed to produce the modernized version of the document.
\end{abstract}

\section{Introduction}
\label{se:intro}
In recent years, awareness of the importance of preserving our cultural heritage has increased. Historical documents are an important part of that heritage. In order to preserve them, there is an increased need in creating digital text versions which can be search and automatically processed \citep{Piotrowski12}. However, their linguistic properties create additional difficulties: due to the lack of a spelling convention, orthography changes depending on the time period and author. Furthermore, human language evolves with the passage of time, increasing the difficulty of the document's comprehension. Thus, historical documents are mostly accessible to scholars specialized in the time period in which each document was written.

Modernization tackles the language barrier in order to increase the accessibility of historical documents. To achieve this, it generates a new version of a historical document in the modern version of the language in which the document was originally written (\cref{fi:Shakespeare} shows an example of modernizing a document). However, while modernization has been successful in order to increase the comprehension of historical documents \citep{Sang17,Domingo18b}, it is still far from creating error-free modern versions. Therefore, this task still needs to be carried out by scholars.

\begin{figure*}[!ht]
	\centering
	\scriptsize
	\begin{minipage}{0.022\textwidth}
		$ $
	\end{minipage}
	\begin{minipage}{0.39\textwidth}
		To be, or not to be? That is the question
		
		Whether 'tis nobler in the mind to suffer
		
		The slings and arrows of outrageous fortune,
		
		Or to take arms against a sea of troubles,
		
		And, by opposing, end them?
		
		To die, to sleep{\textemdash}No more{\textemdash}
		
		and by a sleep to say we end
		
		The heartache and the thousand natural shocks
		
		That flesh is heir to{\textemdash}'tis a consummation
		
		Devoutly to be wished! To die, to sleep.
		
		To sleep, perchance to dream—ay, there’s the rub,
		
		For in that sleep of death what dreams may come
		
		When we have shuffled off this mortal coil,
		
		Must give us pause. There’s the respect
		
		That makes calamity of so long life.
	\end{minipage}
	\begin{minipage}{0.035\textwidth}
		$ $
	\end{minipage}
	\begin{minipage}{0.5\textwidth}
		The question is: is it better to be alive or dead?
		
		Is it nobler to put up with all the nasty things
		
		that luck throws your way,
		
		or to fight against all those troubles
		
		by simply putting an end to them once and for all?
		
		Dying, sleeping{\textemdash}that’s all dying is{\textemdash}
		
		a sleep that ends all
		
		the heartache and shocks
		
		that life on earth gives us{\textemdash}that’s an achievement 
		
		to wish for. To die, to sleep 
		
		{\textemdash}to sleep, maybe to dream. Ah, but there’s the catch:  
		
		in death’s sleep who knows what kind of dreams might come, 
		
		after we’ve put the noise and commotion of life behind us. 
		
		That’s certainly something to worry about. 
		
		That’s the consideration that makes us stretch out our sufferings so long.
	\end{minipage}
	\caption{Example of modernizing a historical document. The original text is a fragment from \emph{Hamlet}. The modernized version of the Sonnet was obtained from \citep{Crowther03}.}
	\label{fi:Shakespeare}
\end{figure*}

Interactive machine translation (IMT) fosters human{\textendash}computer collaborations to generate error-free translations in a productive way \citep{Foster97,Barrachina09}. In this work, we proposed to apply one of these protocols to historical documents modernization. We strive for creating an error-free modern version of a historical document, decreasing the human effort needed to achieve this goal.

The rest of this document is structured as follows: \cref{se:work} introduces the related work. Then, in \cref{se:IMT} we present our protocol. \cref{se:exp} describes the experiments conducted in order to assess our proposal. The results of those experiments are presented and discussed in \cref{se:res}. Finally, in \cref{se:conc}, conclusions are drawn.

\section{Related Work}
\label{se:work}
While the lack of a spelling convention has been extensively researched for years \citep{Baron08,Bollmann16,Domingo18a}, modernization of historical documents is a younger field. \citet{Sang17} organized a shared task in order to translate historical text to contemporary language. The main goal of this shared task was to tackle the spelling problem. However, they also approached document modernization using a set of rules.  \citet{Domingo17a} proposed a modernization approach based on statistical machine translation (SMT). A neural machine translation (NMT) approach was proposed by \citet{Domingo18b}. Finally, \citet{Sen19} extracted parallel phrases from an original parallel corpus and used them as an additional training data for their NMT approach.

Despise the promising results achieved in last years, machine translation (MT) is still far from producing high-quality translations \citep{Dale16}. Therefore, a human agent has to supervise these translation in a post-editing stage. IMT was introduced with the goal of combining the knowledge of a human translator and the efficiency of an MT system. Although many protocols have been proposed in recent years \citep{Marie15,Gonzalez16,Domingo17b,Peris17a}, the prefix-based remains as one of the most successful approaches \citep{Barrachina09,Alabau13,Knowles16}. In this approach, the user corrects the leftmost wrong word from the translation hypothesis, inherently validating a correct prefix. With each new correction, the system generates a suffix that completes the prefix to produce a new translation.

\section{Interactive Machine Translation}
\label{se:IMT}
Classical IMT approaches relay on the statistical formalization of the MT problem. Given a source sentence $\x$, SMT aims at finding its most likely translation $\hat{\y}$~\citep{Brown93}:

\begin{equation}
\hat{\y} = \argmax_{\y} Pr(\y \mid \x)
\label{eq:SMT}
\end{equation}

For years, the prevailing approach to compute this expression have been phrase-based models~\citep{Koehn10}. These models rely on a log-linear combination of different models~\citep{Och02}: namely, phrase-based alignment models, reordering models and language models; among others~\citep{Zens02,Koehn03}. However, more recently, this approach has shifted into neural models (see \cref{se:NMT}).

\subsection{Prefix-based Interactive Machine Translation}
\label{se:PBIMT}
\begin{figure*}[!ht]
	\centering
	\includegraphics[scale=0.75]{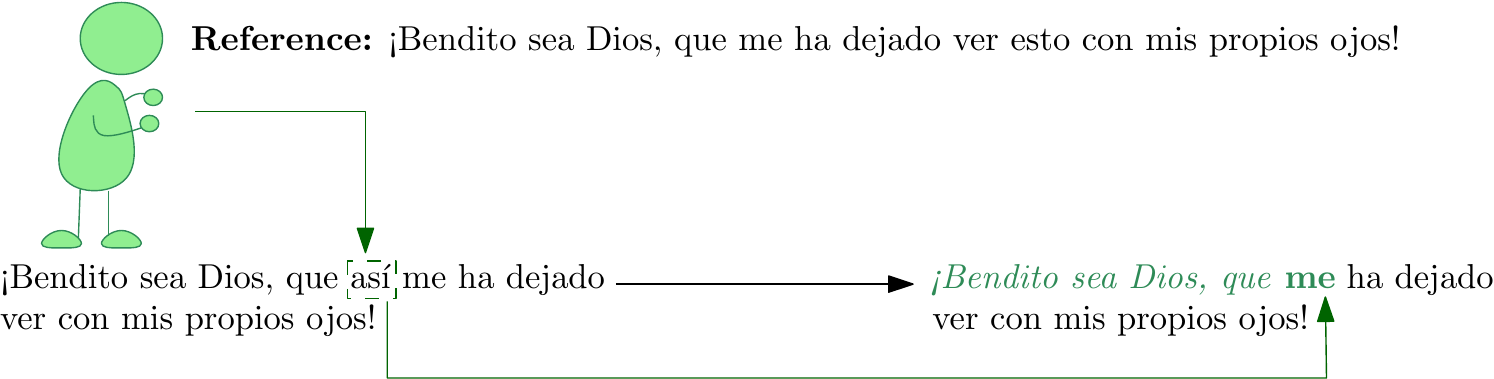}
	\caption{Single iteration of prefix-based IMT. The user corrects the leftmost wrong word \emph{as{\'i}}, introducing the word \textbf{me} at position 5. Then, the system generates a new hypothesis that takes into account the inherently validated prefix (\emph{!`Bendito sea Dios, que \textbf{me}}).}
	\label{fi:IMT}
\end{figure*}

Prefix-based IMT proposed a user{\textendash}computer collaboration that starts with the system proposing an initial translation $\y$ of length $I$. Then, the user corrects the leftmost wrong word $y_i$, inherently validating all preceding words. These words form a validated prefix $\tilde{\y}_p$, that includes the corrected word $\tilde{y}_i$. The system reacts to this user feedback, generating a suffix $\hat{\y}_s$ that completes $\tilde{\y}_p$ to obtain a new translation of $\x:\hat{\y}~=~\tilde{\y}_p\,\hat{\y}_s$. This process is repeated until the user accepts the complete system suggestion. \cref{fi:IMT} illustrates this protocol.

\citet{Barrachina09} formalized the suffix generation as follows:

\begin{equation}
\label{eq:ITP}
\hat{\y}_s = \argmax_{\y_s} Pr(\y_s \mid \x, \tilde{\y}_p)
\end{equation}

\noindent which can be straightforwardly rewritten as:

\begin{equation}
\label{eq:ITPB}
\hat{\y}_s = \argmax_{\y_s} Pr(\tilde{\y}_p\,\y_s \mid \x)
\end{equation}

This equation is very similar to \cref{eq:SMT}: at each iteration, the process consists in a regular search in the translations space but constrained by the prefix $\tilde{\y}_p$.

\subsection{Neural Machine Translation}
\label{se:NMT}
In NMT, \cref{eq:SMT} is modeled by a neural network with parameters $\VTheta$:

\begin{equation}
\hat{\y} \approx \argmax_{\y} \log \p(\y \mid \x; \VTheta)
\label{eq:NMT}
\end{equation}

This neural network usually follows an encoder-decoder architecture, featuring recurrent networks \citep{Bahdanau15,Sutskever14}, convolutional networks \citep{Gehring17} or attention mechanisms \citep{Vaswani17}. Model parameters are jointly estimated on large parallel corpora, using stochastic gradient descent \citep{Robbins51,Rumelhart86}. At decoding time, the system obtains the most likely translation using a beam search method.

\subsection{Prefix-based Interactive Neural Machine Translation}
The prefix-based IMT protocol (see \cref{se:PBIMT}) can be naturally included into NMT systems since sentences are generated from left to right. In order to take into account the user's feedback and generate compatible hypothesis, the search space must be constraint. Given a prefix $\tilde{\y}_p$, only a single path accounts for it. The branching of the search process starts once this path has been covered. Introducing the validated prefix $\tilde{\y}_p$, \cref{eq:NMT} becomes:

\begin{equation}
\label{wq:INMT}
\hat{\y}_s \approx \argmax_{\y_s} \log \p(\y \mid \x; \tilde{\y}_p)
\end{equation}

\noindent which implies a search over the space of translations, but constrained by the validated prefix $\tilde{\y}_p$ \citep{Peris17a}.

\section{Experiments}
\label{se:exp}
In this section, we present our experimental conditions, including translation systems, corpora and evaluation metrics.

\subsection{MT Systems}
SMT systems were trained with \texttt{Moses} \citep{Koehn07}, following the standard procedure: we estimated a 5-gram language model{\textemdash}smoothed with the improved KneserNey method{\textemdash}using \texttt{SRILM} \citep{Stolcke02}, and optimized the weights of the log-linear model with MERT \citep{Och03a}.

We built our NMT systems using \texttt{NMT-Keras} \citep{Peris17b}. We used long short-term memory units \citep{Gers00}, with all model dimensions set to $512$. We trained the system using Adam \citep{Kingma14} with a fixed learning rate of $0.0002$ and a batch size of $60$. We applied label smoothing of $0.1$ \citep{Szegedy15}. At inference time, we used beam search with a beam size of 6. We applied joint byte pair encoding to all corpora \citep{Sennrich16}, using $32,000$ merge operations.

Statistical IMT systems were implemented following the procedure of word graph exploration and generation of a best suffix for a given prefix described by \citet{Barrachina09}. Neural IMT systems were built using the interactive branch of \texttt{NMT-Keras}\footnote{\url{https://github.com/lvapeab/nmt-keras/tree/interactive_NMT}}.

\subsection{Corpora}
\label{se:corp}
The first corpus used in our experimental session was the \textbf{Dutch Bible}~\citep{Sang17}. This corpus consists in a collection of different versions of the Dutch Bible: a version from 1637, another from 1657, another from 1888 and another from 2010. Except for the 2010 version, which is missing the last books, all versions contain the same texts. Moreover, since the authors mentioned that the translation from this last version is not very reliable and, considering that Dutch has not evolved significantly between 1637 and 1657, we decided to only use the 1637 version{\textemdash}considering this as the original document{\textemdash}and the 1888 version{\textemdash}considering 19$^{\mathrm{th}}$ century Dutch as \emph{modern Dutch}.

We selected \textbf{El Quijote} \citep{Domingo18b} as our second corpus. This corpus contains the famous 17$^{\mathrm{th}}$ century Spanish novel by Miguel de Cervantes, and its correspondent 21$^{\mathrm{st}}$ century version. 

Finally, we used \textbf{El Conde Lucanor} \citep{Domingo18b} as a third corpus. This data set contains the original 14$^{\mathrm{th}}$ century Spanish novel by Don Juan Manuel, and its correspondent 21$^{\mathrm{st}}$ century version. 
Due to the small size of the corpus, we decided to use it only as a test. Additionally, unable to find a suitable training corpus, we used the systems built for \emph{El Quijote}{\textemdash}despite the original documents belonging to different time periods{\textemdash}in order to modernize \emph{El Conde Lucanor}.

\cref{ta:corp} presents the corpora statistics.

\begin{table*}[!ht]
	\centering
	\resizebox{0.9\textwidth}{!}{\begin{minipage}{\textwidth}
			\centering
			\begin{tabular}{c c c c c}
				\toprule
				&  & \textbf{Dutch Bible} & \textbf{El Quijote} & \textbf{El Conde Lucanor} \\
				\midrule
				\multirow{3}{*}{Train} & $|S|$ & 35.2K & 10K & - \\
				& $|T|$ & 870.4/862.4K & 283.3/283.2K & - \\
				& $|V|$ & 53.8/42.8K & 31.7/31.3K & - \\
				\midrule
				\multirow{3}{*}{Development} & $|S|$ & 2000 & 2000 & - \\
				& $|T|$ & 56.4/54.8K & 53.2/53.2K & - \\
				& $|V|$ & 9.1/7.8K & 10.7/10.6K & - \\
				\midrule
				\multirow{3}{*}{Test} & $|S|$ & 5000 & 2000 & 2252 \\
				& $|T|$ & 145.8/140.8K & 41.8/42.0K & 62.0/56.7K \\
				& $|V|$ & 10.5/9.0K & 8.9/9.0K & 7.4/8.6K \\
				\bottomrule
			\end{tabular}
	\end{minipage}}
	\caption{Corpora statistics. $|S|$ stands for number of sentences, $|T|$ for number of tokens and $|V|$ for size of the vocabulary. \emph{Monolingual} refers to the monolingual data used to create the synthetic data. M denotes million and K thousand.}
	\label{ta:corp}
\end{table*}


\subsection{Metrics}
In order to measure the gains in human effort reduction, we made use of the following metrics:

\begin{description}
	\item[Word Stroke Ratio (WSR)] \citep{Tomas06}: measures the number of words edited by the user, normalized by the number of words in the final translation.
	
	\item[Mouse Action Ratio (MAR)] \citep{Barrachina09}: measures the number of mouse actions made by the user, normalized by the number of characters in the final translation.
\end{description}

Additionally, to evaluate the quality of the modernization and the difficulty of each task, we made use of the following well-known metrics:

\begin{description}
	\item \textbf{BiLingual Evaluation Understudy (BLEU)} \citep{Papineni02}: computes the geometric average of the modified n-gram precision, multiplied by a brevity factor that penalizes short sentences.
	\item \textbf{Translation Error Rate (TER)} \citep{Snover06}: computes the number of word edit operations (insertion, substitution, deletion and swapping), normalized by the number of words in the final translation.
\end{description}

We used sacreBLEU \citep{Post18} for ensuring consistent BLEU scores. For determining whether two systems presented statistically significant differences, we applied approximate randomization tests \citep{Riezler05}, with $10,000$ repetitions and using a $p$-value of $0.05$.

\subsection{User Simulation}
Due to the high costs of an evaluation involving human agents, we carried out an automatic evaluation with simulated users whose desired modernizations correspond to the reference sentences.

At each iteration, the user corrects the leftmost wrong word from the system's hypothesis. With this correction, a new prefix is validated. The associated cost of this correction is of one mouse action and one word stroke. The system, then, reacts to this feedback, generating a new suffix that completes the prefix to conform a new hypothesis. This process is repeated until hypothesis and reference are the same.

\section{Results}
\label{se:res}
\cref{ta:quality} presents the quality of the modernization. Both SMT and NMT approaches were able to significantly improved the baseline. That is, the modernized documents are easier to comprehend by a contemporary reader than the original documents. An exception to this is \emph{El Conde Lucanor}. The SMT approach yielded significant improvements in terms of TER, but was worse in terms of BLEU. Moreover, the NMT approach yielded worst results in terms of both BLEU and TER. Most likely, this results are due to having used the systems trained with \emph{El Quijote} for modernizing \emph{El Conde Lucanor} (see \cref{se:corp}).

\begin{table*}[!ht]
	\centering
	\resizebox{0.9\textwidth}{!}{\begin{minipage}{\textwidth}
			\centering
			\begin{tabular}{c c c c c c c}
				\toprule
				\multirow{2}{*}{\textbf{System}} &  \multicolumn{2}{c}{\textbf{Dutch Bible}} & \multicolumn{2}{c}{\textbf{El Quijote}} & \multicolumn{2}{c}{\textbf{El Conde Lucanor}} \\
				\cmidrule(lr){2-3}\cmidrule(lr){4-5}\cmidrule(lr){6-7}
				& BLEU [$\uparrow$] & TER [$\downarrow$] & BLEU [$\uparrow$] & TER [$\downarrow$] & BLEU [$\uparrow$] & TER [$\downarrow$] \\
				\midrule
				Baseline & $13.5$ & $57.4$ & $36.2$ & $48.6$ & $\mathbf{7.9}$ & $90.5$ \\
				\midrule
				SMT & $\mathbf{77.6^{\dagger\ddagger}}$ & $\mathbf{11.6^{\dagger\ddagger}}$ & $\mathbf{57.1^{\dagger\ddagger}}$ & $\mathbf{30.7^{\dagger\ddagger}}$ & $7.4^{\dagger\ddagger}$ & $\mathbf{84.6^{\dagger\ddagger}}$ \\
				NMT & $69.2^\dagger$ & $18.1^\dagger$ & $39.7^\dagger$ & $52.5^\dagger$ & $4.1^\dagger$ & $98.1^\dagger$ \\
				\bottomrule
			\end{tabular}
	\end{minipage}}
	\caption{Modernization quality. \emph{Baseline} system corresponds to considering the original document as the modernized version. \emph{SMT} and \emph{NMT} are the SMT and NMT approaches respectively. $^\dagger$ indicates statistically significant differences between the SMT/NMT system and the baseline. $^\ddagger$ indicates statistically significance between the NMT and SMT systems. Best results are denoted in bold.}
	\label{ta:quality}
\end{table*}

When comparing the SMT and NMT approaches, we observe that SMT yielded the best results in all cases. This behavior was already perceived by  \citet{Domingo18b} and is, most likely, due to the small size of the training corpora{\textemdash}a well-known problem in NMT. However, while the goal of modernization is making historical documents as easier to comprehend by contemporary people as possible, our goal is different. In this work, our goal is to obtain an error-free modern copy of a historical document. To achieve this, we proposed an interactive collaboration between a human expert and our modernizing system, in order to reduce the effort needed to generate such copy. \cref{ta:effort} presents the experimental results.

\begin{table*}[!ht]
	\centering
	\resizebox{0.9\textwidth}{!}{\begin{minipage}{\textwidth}
			\centering
			\begin{tabular}{c c c c c c c}
				\toprule
				\multirow{2}{*}{\textbf{System}} &  \multicolumn{2}{c}{\textbf{Dutch Bible}} & \multicolumn{2}{c}{\textbf{El Quijote}} & \multicolumn{2}{c}{\textbf{El Conde Lucanor}} \\
				\cmidrule(lr){2-3}\cmidrule(lr){4-5}\cmidrule(lr){6-7}
				& WSR [$\downarrow$] & MAR [$\downarrow$] & WSR [$\downarrow$] & MAR [$\downarrow$] & WSR [$\downarrow$] & MAR [$\downarrow$] \\
				\midrule
				Baseline & $63.4$ & $13.3$ & $46.5$ & $11.9$ & $\mathbf{85.9}$ & $\mathbf{21.3}$ \\
				\midrule
				SMT & $\mathbf{14.8^{\dagger\ddagger}}$ & $\mathbf{4.5^{\dagger\ddagger}}$ & $\mathbf{39.7^{\dagger\ddagger}}$ & $\mathbf{10.8^{\dagger\ddagger}}$ & $95.4^{\dagger\ddagger}$ & $24.1^{\dagger\ddagger}$ \\
				NMT & $28.9^\dagger$ & $6.2^\dagger$ & $62.3^\dagger$ & $13.0^\dagger$ & $96.2^\dagger$ & $19.5^\dagger$ \\
				\bottomrule
			\end{tabular}
	\end{minipage}}
	\caption{IMT results. \emph{SMT} and \emph{NMT} are the IMT approaches based on SMT and NMT respectively. $^\dagger$ indicates statistically significant differences between the SMT/NMT system and the baseline. $^\ddagger$ indicates statistically significance between the NMT and SMT systems. Best results are denoted in bold.}
	\label{ta:effort}
\end{table*}

Both SMT and NMT approaches yielded significant reductions of the human effort needed to modernize the \emph{Dutch Bible} (up to $48$ points in terms of WSR and $8$ in terms of MAR) and \emph{El Quijote} (up to $7$ points in terms of WSR and $1$ of MAR). For \emph{El Conde Lucanor}, however, both approaches resulted in an increased of the effort need to generate an error-free modern version. This behavior was to be expected since the modernization quality for \emph{El Conde Lucanor} was very low. Therefore, the system consistently generated wrong suffixes, resulting in the user having to make more corrections.

Regarding the performance of both approaches, SMT achieved the highest effort reduction. This was reasonably expected since its modernization quality was better. However, in past neural IMT works \citep{Peris17a}, the neural IMT approach was able to yield further improvements despite having a lower translation quality than its SMT counterpart. Most likely, the reason of this is that, due to the small training corpora, the neural model was not able to reach its best performance, Nonetheless, we should address this in a future work.

\subsection{Qualitative Analysis}
\cref{fi:exIMT} shows an example of modernizing a sentence from \emph{El Quijote} with the interactive SMT approach. While the system's initial suggestion contains five errors, with the IMT protocol, the user only needs to make three corrections. With each correction, the system is able to improve its suggestions, reducing the total effort needed to achieve an error-free modernization. Note that this example has been chosen for illustrative purposes of a correct functioning of the system. The average sentences from \emph{El Quijote} are longer, and there are times in which the system fails to take the human knowledge into account, resulting in an increase of the number of corrections. Nonetheless, as seen in \cref{se:res}, overall the system is able to significantly decrease the human effort.

\begin{figure*}[!ht]
	\centering
	\bgroup
	\def\arraystretch{1.25}
	\resizebox{0.9\textwidth}{!}{\begin{minipage}{\textwidth}
			\centering
			\begin{tabular}{c | c | l}
				\multicolumn{3}{l}{\textbf{source ($\bf{x}$):} durmamos por aora entrambos, y despues, Dios dixo lo que sera.} \\
				\multicolumn{3}{l}{\textbf{target translation ($\bf{\hat{y}}$):} Durmamos de momento los dos, y despu{\'e}s Dios dir{\'a}.} \\
				\hline
				\textbf{IT-0} & $\bf{MT}$ &  Durmamos por ahora ambos, y despu{\'e}s Dios dir{\'a}. \\
				\hline
				\multirow{2}{*}{\textbf{IT-1}} & $\bf{User}$ & \regularbox{Durmamos} \textbf{de} ahora ambos, y despu{\'e}s Dios dir{\'a}. \\
				& $\bf{MT}$ & \regularbox{Durmamos de} momento ambos, y despu{\'e}s Dios dir{\'a}. \\
				\hline
				\multirow{2}{*}{\textbf{IT-2}} & $\bf{User}$ & \regularbox{Durmamos de momento} \textbf{los} y despu{\'e}s Dios dir{\'a}. \\
				& $\bf{MT}$ & \regularbox{Durmamos de momento los} dos y despu{\'e}s Dios dir{\'a}. \\
				\hline
				\textbf{END} & $\bf{User}$ & Durmamos de momento los dos, y despu{\'e}s Dios dir{\'a}. \\
				\hline
			\end{tabular}
	\end{minipage}}
	\egroup
	\caption{IMT session to modernize a sentence from \emph{El Quijote}. At the initial iteration (\emph{IT-0}), the system suggests an initial modernization. Then, at iteration 1, the user corrects the leftmost wrong word ({\footnotesize\textbf{de}}). With this action, the user is inherently validating the prefix {\footnotesize\regularbox{Durmamos}}. Taking this user feedback into account, the system suggests a new hypothesis. Similarly, at iteration 2, the user corrects the leftmost wrong word ({\footnotesize\textbf{los}}). The session ends when the user accepts the last modernization suggested by the system.}
	\label{fi:exIMT}
\end{figure*}

\begin{figure*}[!ht]
	\centering
	\bgroup
	\def\arraystretch{1.25}
	\resizebox{0.9\textwidth}{!}{\begin{minipage}{\textwidth}
			\centering
			\begin{tabular}{c | c | l}
				\multicolumn{3}{l}{\textbf{source ($\bf{x}$):} durmamos por aora entrambos, y despues, Dios dixo lo que sera.} \\
				\multicolumn{3}{l}{\textbf{target translation ($\bf{\hat{y}}$):} Durmamos de momento los dos, y despu{\'e}s Dios dir{\'a}.} \\
				\hline
				\textbf{IT-0} & $\bf{MT}$ &  {\textemdash}durdos por ahora los amores, y despu{\'e}s, Dios dijo all{\'i}. \\
				\hline
				\multirow{2}{*}{\textbf{IT-1}} & $\bf{User}$ & \textbf{Durmamos} por ahora los amores, y despu{\'e}s, Dios dijo all{\'i}. \\
				& $\bf{MT}$ & \regularbox{Durmamos} duradas por ahora los amores , y después , Dios dijo allí . \\
				\hline
				\multirow{2}{*}{\textbf{IT-2}} & $\bf{User}$ & \regularbox{Durmamos} \textbf{de} por ahora los amores, y despu{\'e}s, Dios dijo all{\'i}. \\
				& $\bf{MT}$ & \regularbox{Durmamos de} pasar por ahora los buenos, y , si Dios quiere. \\
				\hline
				\multirow{2}{*}{\textbf{IT-3}} & $\bf{User}$ & \regularbox{Durmamos de} \textbf{momento} por ahora los buenos, y , si Dios quiere. \\
				& $\bf{MT}$ & \regularbox{Durmamos de momento}  en que los mante{\'o}, y despu{\'e}s, Dios dijo all{\'i}. \\
				\hline
				\multicolumn{3}{c}{...} \\
				\hline
				\multirow{2}{*}{\textbf{IT-8}} & $\bf{User}$ & \regularbox{Durmamos de momento los dos, y despu{\'e}s Dios} \textbf{dir{\'a}.}  \\
				& $\bf{MT}$ & \regularbox{Durmamos de momento los dos, y despu{\'e}s Dios dir{\'a}.}  \\
				\hline
				\textbf{END} & $\bf{User}$ & Durmamos de momento los dos, y despu{\'e}s Dios dir{\'a}. \\
				\hline
			\end{tabular}
	\end{minipage}}
	\egroup
	\caption{Neural IMT session to modernize the same sentence from \emph{El Quijote} as in \cref{fi:exIMT}. At the initial iteration (\emph{IT-0}), the system suggests an initial modernization. Then, at iteration 1, the user corrects the leftmost wrong word ({\footnotesize\textbf{Durmamos}}). Taking this user feedback into account, the system suggests a new hypothesis. Similarly, at iteration 2, the user corrects the leftmost wrong word ({\footnotesize\textbf{de}}). The session ends when the user accepts the last modernization suggested by the system.}
	\label{fi:exINMT}
\end{figure*}

\cref{fi:exINMT} contains an example of modernizing the same sentence as in \cref{fi:exIMT}, using the interactive NMT approach. This is an example in which the system fails to take into account the user's corrections, resulting in an increase of the human effort. It is specially worth noting the introduction of non-existing words such as \emph{durdos} and \emph{duradas}. This problem was probably caused by an incorrect segmentation of a word, via the byte pair encoding process, and should be address in a future work. Nonetheless, as seen in \cref{se:res}, overall the system is able to significantly decrease the human effort.

\section{Conclusions and Future Work}
\label{se:conc}
In this work, we proposed a collaborative user{\textendash}computer approach to create an error-free modern version of a historical document. We tested this proposal on a simulated environment, achieving significant reductions of the human effort. We built our modernization protocol based on both SMT and NMT approaches to prefix-based IMT. Although both systems yielded significant improvements for two data sets out of three, the SMT approach yielded the best results{\textemdash}both in terms of the human reduction and in the modernization quality of the initial system.

As a future work, we want to further research the behavior of the neural systems. For that, we would like to explore techniques for enriching the training corpus with additional data, and the incorrect generation of words due to subwords. We would also like to develop new protocols based on successful IMT approaches. Finally, we should test our proposal with real users to obtain actual measures of the effort reduction.

\section*{Acknowledgments}
The research leading to these results has received funding from the European Union through \emph{Programa Operativo del Fondo Europeo de Desarrollo Regional (FEDER)} from Comunitat Valencia (2014{\textendash}2020) under project \emph{Sistemas de frabricación inteligentes para la indústria 4.0} (grant agreement IDIFEDER/2018/025); and from  Ministerio de Econom{\'i}a y Competitividad (MINECO) under project \emph{MISMIS-FAKEnHATE} (grant agreement PGC2018-096212-B-C31). We gratefully acknowledge the support of NVIDIA Corporation with the donation of a GPU used for part of this research.

\bibliographystyle{apalike}
\bibliography{tex/IMT}

\end{document}